\documentclass[journal]{IEEEtran}

\usepackage{graphicx}

\usepackage{amsmath}
\usepackage{bm}
\usepackage{amssymb}
\usepackage{gensymb}
\usepackage{subfig}

\usepackage{algorithm}
\usepackage{algpseudocode}

\usepackage{hyperref}
\hypersetup{
  colorlinks=true,
  urlcolor=green,
  citecolor=green
}

\ifCLASSINFOpdf
\else
\fi

\begin{document}

\title{Semi-autonomous Prosthesis Control Using Minimal Depth Information and Vibrotactile Feedback}

\author{Miguel~Nobre~Castro
        and~Strahinja~Dosen
\thanks{Received 18 February 2025; revised 8 June 2025; accepted 12 July 2025. This study was supported by Independent Research Fund Denmark through the projects ROBIN (8022-00243A) and CLIMB (2035-00169A).
(\textit{Corresponding author: Strahinja Dosen}).
}%
\thanks{The authors are with the Neurorehabilitation Systems Research Group, Department
of Health Science and Technology, Faculty of Medicine, Aalborg University, Aalborg, Denmark (e-mail: sdosen@hst.aau.dk).}%
}

\maketitle

\begin{abstract}
Semi-autonomous prosthesis controllers based on computer vision improve performance while reducing cognitive effort.
However, controllers relying on full-depth data face challenges in being deployed as embedded prosthesis controllers due to the computational demands of processing point clouds.
To address this, the present study proposes a method to reconstruct the shape of various daily objects from minimal depth data.
This is achieved using four concurrent laser scanner lines instead of a full point cloud.
These lines represent the partial contours of an object's cross-section, enabling its dimensions and orientation to be reconstructed using simple geometry.
A control prototype was implemented using a depth sensor with four laser scanners.
Vibrotactile feedback was also designed to help users to correctly aim the sensor at target objects.
Ten able-bodied volunteers used a prosthesis equipped with the novel controller to grasp ten objects of varying shapes, sizes, and orientations.
For comparison, they also tested an existing benchmark controller that used full-depth information.
The results showed that the novel controller handled all objects and, while performance improved with training, it remained slightly below that of the benchmark.
This marks an important step towards a compact vision-based system for embedded depth sensing in prosthesis grasping.
\end{abstract}


%
\IEEEpeerreviewmaketitle

\section{Introduction}
\IEEEPARstart{M}{odern} advanced hand prostheses are still difficult to operate, leading to lower adoption rates among users \cite{Salminger2020}.
Myoelectric signals recorded by their embedded sensors enable to estimate user intent and translate it into prosthesis commands.
Despite this direct and intuitive user-prosthesis connection, the subject is responsible for explicitly controlling all system functions.
That can be slow and cognitively taxing, especially in the case of complex systems with many degrees of freedom.
Myoelectric control methods that use machine learning hold promise for improving such interactions \cite{marinelli2023active}, but those that reached the commercial stage rely on sequential control as the user can activate only one function at a time \cite{myoplus_url, coapt_url}.

An alternative approach to address this challenge is to make the prosthesis semi-autonomous.
In this case, the device has additional sensors and automatic functions, which the user can trigger using a simple myoelectric interface \cite{gvargas2015hmi}.
Upon triggering, the prosthesis can be programmed to follow: the trajectory of the upper extremity \cite{gloumakov2022trajectory}; the movement of the contra-lateral hand \cite{volkmar2019bimanual}; or even to react to a scene perceived by a vision sensor, allowing the artificial controller to “see”, for instance, an object that the user wants to grasp \cite{dovsen2010cognitive}.

In the latter case, the controller can infer object properties (shape, size, and orientation) from the imaging data.
This information can then be used to preshape the hand automatically by simultaneously activating multiple degrees-of-freedom (DoF) \cite{castro2022continuous}. Since the controller performs these functions automatically, it can alleviate the cognitive burden, and the tests have shown that it can outperform classification-based approaches \cite{mouchoux2021decreases}.

\begin{figure}[t!]
\centerline{\includegraphics[width=3.5in]{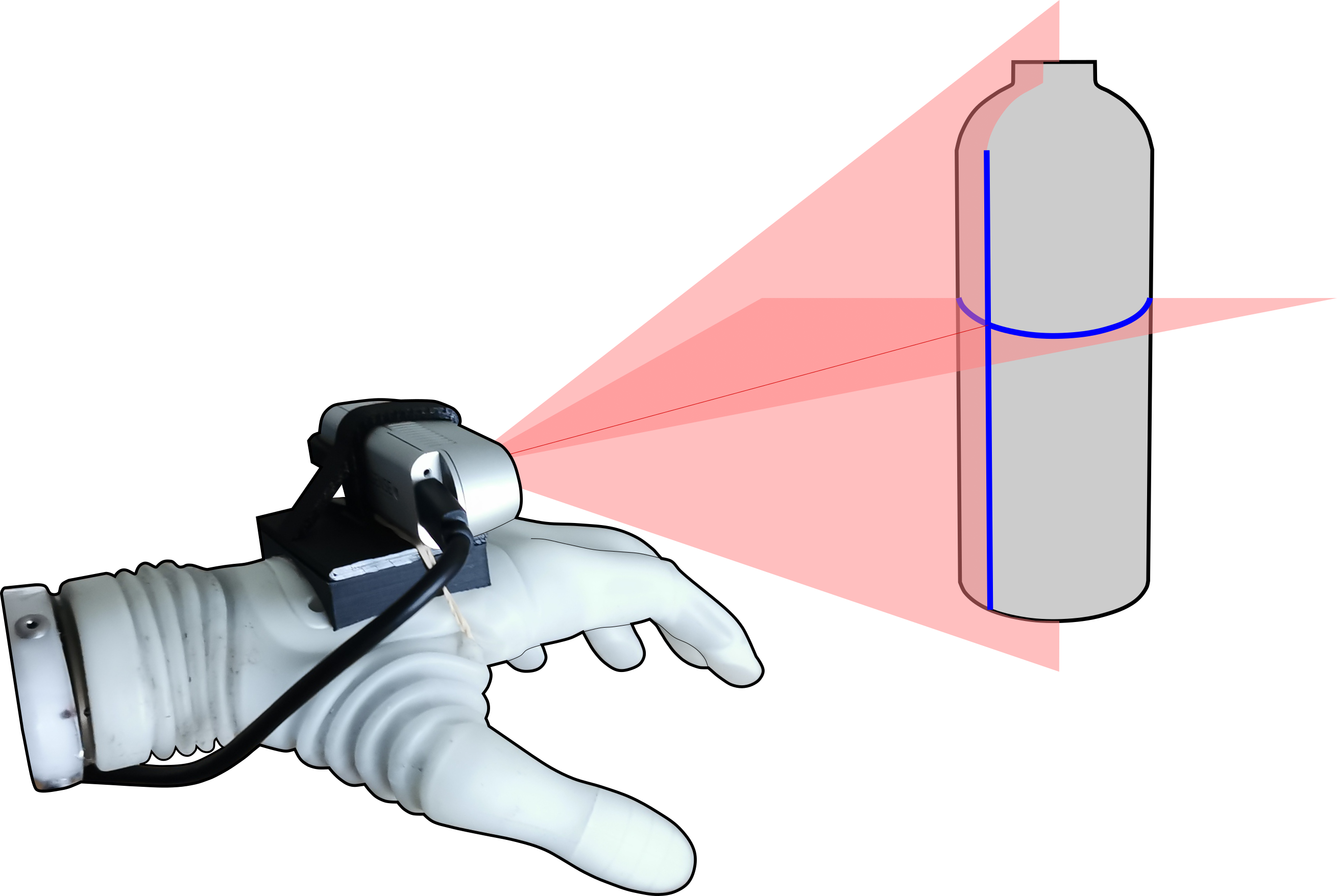}}
\caption{Conceptual illustration of the semi-autonomous prosthesis control based on minimal data. Multiple 2-D laser scanners (red planes) can yield multiple laser “cuts” throughout an object. Each of these scan lines can be fitted by a 2-D model (blue lines — circle, straight line or ellipse) and used to estimate object properties which allow to automatically preshape the prosthesis.}
\label{fig:contour}
\end{figure}

Different types of vision sensors have been used to implement semi-autonomous control: regular webcams \cite{dovsen2010cognitive,ghazaei2017deep,gardner2020multimodal,zhong2020reliable,vasile2022synthetic}, depth sensors \cite{castro2022continuous,markovic2015sensor}, and even augmented reality (AR) headsets \cite{markovic2014stereo, mouchoux2021decreases}.
The sensors were mostly placed on the head (egocentric) \cite{markovic2014stereo,markovic2015sensor,gardner2020multimodal,mouchoux2021decreases} or the prosthesis, either on the prosthetic hand \cite{dovsen2010cognitive,castro2022continuous} or the forearm \cite{zhong2020reliable,vasile2022synthetic}.
In addition, recent studies presented solutions in which small RGB cameras were integrated into the hand \cite{starke2022inpalm,castro2022hybrid} or fingers \cite{hundhausen2021infinger}.
Placing the sensor on the hand enables a self-contained system, but it also requires an aiming step, where the user aligns the prosthesis towards a given object so that the object is in the sensor's field-of-view \cite{castro2022continuous}.

A disadvantage of the vision-based semi-autonomous approaches is that they rely on heavy data processing.
For instance, RGB-based controllers use classic image processing, such as blob analysis \cite{dovsen2010cognitive} or edge detection \cite{gardner2020multimodal}, or deep neural networks \cite{ghazaei2017deep,shi2020pattern,zhong2020reliable,vasile2022synthetic}.
Some of these approaches even process multiple frames in a grasping sequence to capture the temporal and dynamic nature of the action \cite{zhong2020reliable,vasile2022synthetic}.

Likewise, depth-based controllers reconstruct object shapes from 3-D point cloud data, which is more computationally intensive and involve the fitting of generic geometric 3-D models (e.g. a sphere, cylinder or cuboid) to the data.
Often, this procedure includes a preprocessing step of surface/planar data removal \cite{markovic2014stereo,markovic2015sensor} or segmentation \cite{mouchoux2021decreases,castro2022continuous}.
A recent study \cite{castro2022continuous} showed that depth sensing can be employed in continuous semi-autonomous control that can be readily used for grasping after brief training.
A general overview of methods for shared control of hand prostheses using computer vision and other sensing technologies is presented in \cite{guo2023toward}.

However, the computational complexity of estimating object properties using depth sensing and other computer vision approaches poses a critical challenge.
This is because successful clinical translation requires the processing pipeline to run embedded on a prosthesis with limited hardware resources.
Less data and simpler algorithms could potentially reduce such complexity, ideally without compromising performance.
While depth sensors can retrieve the 3-D structure of an entire scene — i.e., the environment —, most of that data is not relevant for prosthesis grasping; the critical information is the structure of the object itself.
In this sense, 2-D laser scanners can retrieve minimal but still relevant depth data.

Laser scanners have been already applied to enhance grasping in dexterous robotic hands.
The DLR hand \cite{Butterfass1998} included embedded light projection laser diodes at the tip of each finger to support reconstruction using a passive stereo-camera pair.
Many years later, Quigley et al. \cite{Quigley2014} similarly used a laser line scanner on the back of one of the robotic fingertips of their dexterous hand, also for object reconstruction through multiple finger sweeps.
Still, this technology has not yet been applied to the control of grasping in semi-autonomous prostheses.
This context is very different from conventional robotics because the user decides what to grasp, and then transports and orients the prosthesis, while the controller needs to dynamically react and adapt to such user behavior.

The present manuscript investigates whether a set of laser scanners can be used to fully reconstruct the shape, size, and orientation of different objects.
While many state-of-the-art vision-based methods for prosthesis control use full depth information \cite{markovic2014stereo,markovic2015sensor,mouchoux2021decreases,castro2022continuous}, this work explores the idea of performing multiple laser “cuts” capable of providing partial contours of an object's cross-sections and to reconstruct the object properties from these contours using simple geometry.
To test this concept functionally, a semi-autonomous prosthesis controller was implemented and assessed on ten able-bodied volunteers who used the prosthesis to perform a functional task.
To develop and demonstrate the method, we used a depth sensor to acquire a full point cloud of the object, from which we extracted the relevant scan lines, thereby flexibly emulating the real laser scanner.
Embedding a set of laser scanners into the prosthetic hand to emit the rays at specific angles is a separate technical and implementation challenge, and hence outside the scope of the present work, which focused on presenting and assessing the novel method.
The placement of the sensors on the prosthesis hand requires the user to aim the prosthesis towards the object, which is not trivial and can take time.
To facilitate aiming, we have further equipped the prosthesis with a vibrotactile system that provides real-time aiming cues to help properly orient the prosthesis.
Whereas vibrotactile stimulation is commonly used to provide feedback about the prosthesis state (e.g., grasping force) \cite{marinelli2023active}, the use of such stimulation to assist object targeting in a semi-autonomous prosthesis is an original contribution of the present study.

Finally, the novel system's performance was tested against a recently developed benchmark \cite{castro2022continuous}, which entails more complex processing since it uses the full depth data rather than a few “discrete” scans as in the novel prototype.
Our hypothesis was that, after some training, the minimal system plus vibrotactile feedback would perform on par with the benchmark system while using substantially less data processing.
To measure the performance, we assessed the accuracy of object reconstruction as well as the time to accomplish the task.
Moreover, the novel system was tested against irregularly shaped objects to demonstrate its robustness.
In summary, this work aimed to demonstrate the feasibility of object reconstruction using minimal data, establish limitations in the number of laser scans, develop methods, and translate them into a functional semi-autonomous controller prototype.

\section{Methods}

\subsection{Reconstruction of 3-D Objects Using Laser Scanners}
\label{sec:laser}
A hypothetical laser scanner using four concurrent laser lines spaced 45\degree~apart was designed for this study, since it allows estimating all important information for secure grasping.
As later explained (Sec. \ref{sec:geometric}), this configuration enables accurate estimation of both dimensions and orientation for all object shapes considered, using simple geometry.
More details on the minimum number of laser lines needed to reconstruct basic 3-D geometric shapes can be found in the Supplementary Material (Sec S.I).

\subsubsection{Geometric Reconstruction of 3-D Models} \label{sec:geometric}

\begin{figure*}[t!]
\centerline{\includegraphics[width=5in]{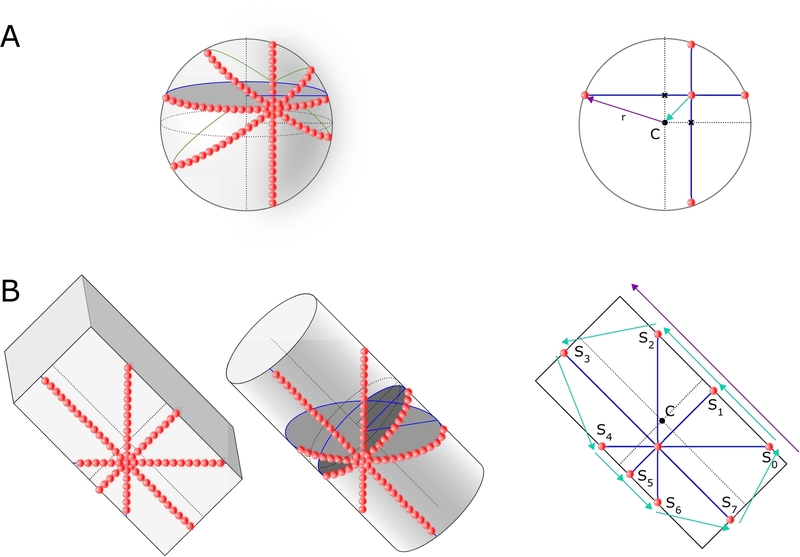}}
\caption{The reconstruction of 3-D models from 2-D models.
(A) The radius $r$ and center $C$ of a spherical object can be retrieved using just two scans by fitting two circle models — the black dots indicate the centers of the two fitted circles. (B) Cuboid and cylindrical objects require four scans: four straight lines if cuboid; four ellipses or two ellipses, one circle, and a straight line if cylindrical — eight red seed points $S_{i}$ points obtained from the scans allow to find the cross-sectional plane of the object, as well as its longitudinal direction vector (purple). More details provided in the text.}
\label{fig:geometric}
\end{figure*}

Fig. \ref{fig:geometric} illustrates how shape, dimension, and orientation of spherical, cylindrical, or cuboid objects can be determined using four scan lines.
The line where all laser scans intersect is referred to as the optical $z$-axis.
Spherical objects (Fig. \ref{fig:geometric}-A) are a special case, as only two laser scans are required.
Each scan results in a partial circular contour, allowing a circle model to be fitted to the contour data using Random Sample Consensus (RANSAC) \cite{fischler1981random}.
The centers of the fitted circle models lie in a plane perpendicular to the optical $z$-axis that contains the center of the sphere.
Therefore, the $z$-coordinate of that plane, and hence the $z$-coordinate of the sphere's center, can be estimated by averaging the $z$-coordinates of the two circles.
The $x$- and $y$-coordinates of the sphere's center (point $C$ in Fig. \ref{fig:geometric}-A, right) are directly obtained from the $x$- and $y$-coordinates of the two circle centers, respectively.
Finally, the radius of the sphere can be calculated as the vector distance between the sphere's center point $C$ and one of the extreme points.

The reconstruction of cuboid and cylindrical objects is more complex, as it requires fitting 2-D geometric models using all four laser scans (see Fig. \ref{fig:geometric}-B).
For a cuboid, four line models need to be fitted, while a cylinder may be represented by four ellipses, or by two ellipses, a circle, and a line, depending on the position and orientation of the optical $z$-axis (i.e., how the user aims at the object).
The first step is to identify the plane corresponding to either the face of the cuboid “pierced" by the optical $z$-axis or, in the case of a cylinder, the longitudinal plane that divides it into two semi-cylinders.
To achieve this, the extreme “inlier" points of each laser scan are found.
For the cuboid, the face plane is determined by calculating the cross product between two vectors formed from three non-collinear extreme points.
In the case of the cylinder, the longitudinal plane is similarly calculated, with the additional constraint that the extreme points must belong to a circle or an ellipse.
The center of the cylinder is then estimated by averaging the coordinates of the centers of the circles and/or ellipses.
Once these planes are determined, the orientation and dimensions of each 3-D objects is calculated using the algorithm illustrated in Fig. \ref{fig:geometric}-B, right.

First, all extreme points are projected onto their respective plane.
The projected points are designated as seed points $S$ and sorted counterclockwise $(S_{0}$ to $S_{7})$ as in Fig. \ref{fig:geometric}-B, right.
At each seed point $S_{i}$, if the measured angle between the inbound vector $(S_{i}-S_{i-1})$ and outbound vector $(S_{i+1}-S_{i})$ is less than $\alpha_{max}$ (i.e., the vectors are nearly parallel), then the vector $(S_{i+1}-S_{i-1})$ is one of the principal directions of the shape.
The pseudocode for this process is presented in the Algorithm \ref{alg:seeds} below.

\begin{algorithm}
\caption{
Finds one of the principal directions of a cuboid or cylindrical object,
where $N$ represents the number of seed points $S$ and $\alpha_{max}$ the maximum threshold angle.
}\label{alg:seeds}
\begin{algorithmic}
\Require $N \gets 8$
\State $\alpha_{max} \gets 10\degree$
\State $i \gets 0$
\While{$i < N$}
\If{$\measuredangle(S_{i}-S_{i-1}, S_{i+1}-S_{i}) \leq \alpha_{max}$}
\Return $S_{i+1}-S_{i-1}$
\EndIf
\State $i \gets i + 1$
\EndWhile
\end{algorithmic}
\end{algorithm}

To identify the shape of a given target object (spherical, cuboid, or cylindrical), three 2-D models (circle, line, and ellipse) are concurrently fitted to the data from each laser scan.
The count of the best-fitted models is then used to determine the reconstructed object shape:
i) if all models are circles, the object is classified as a sphere;
ii) if all models are lines, the object is classified as a cuboid;
iii) if the models include all ellipses, or a combination of one circle, one line, and two ellipses, the object is classified as a cylinder.
For computational efficiency, the RANSAC fitting of the 2-D models should be implemented in parallel (through multi-threading).

Given that a circle is a particular case of an ellipse with equal semi-axis lengths, a rule is imposed such that a cylindrical model is chosen only when its fitting percentage is more than twice that of the spherical model.
Even though no 2-D circle models are used to fit the cylindrical 3-D model during reconstruction, this model requires two RANSAC fittings (of an ellipse and a straight line) per laser line, yielding a total of eight RANSAC fittings for reconstructing a cylindrical model.
Thus, the most time-consuming operation comes from the system trying to fit a cylindrical model to any given object.
Nevertheless, checking how each 3-D model shape fits the data allows for improving the robustness of the system by ensuring the best reading of an object's principal axis and diameter, which are crucial for estimating wrist orientation and grasp aperture.

\subsection{Semi-autonomous Controller Prototype} \label{sec:control}

\subsubsection{System Operation}  \label{sec:sysoperation}

\begin{figure}[t!]
\centerline{\includegraphics[width=\columnwidth]{figures/control_schematic.jpg}}
\caption{The schematic of the semi-autonomous prosthesis control approach with feedback-assisted aiming. The autonomous controller is triggered by a user command to analyze the laser scans and preshape the prosthesis to a suitable configuration for grasping an object. Vibrotactile feedback controlled by a depth processing pipeline constantly cues and guides the user to aim with the prosthesis within the object — a precondition for a successful reconstruction. At any time, the user can take over the control of the prosthesis (direct control) and then move it using proportional myoelectric commands.}
\label{fig:control}
\end{figure}

When operating the actual prosthesis, the vision-based semi-autonomous control used in this study combines automatic and manual/direct control (Fig. \ref{fig:control}).
A two-channel myoelectric interface, similar to the one in \cite{castro2022continuous}, enables the user to switch between these two control modes.
Importantly, manual control is implemented as the standard commercial approach to prosthesis control \cite{vujaklija2016new}.
The activation of wrist and hand flexor and extensor muscles proportionally moves a single degree of freedom (e.g., closes and opens the hand), while the co-activation of both muscle groups, switches the active degree of freedom (e.g., from opening/closing to controlling rotation).

With the prosthesis in idle state, data points within the capture volume are collected while waiting for the user to generate a simple myoelectric command to trigger automatic control.
The user is responsible for aiming at the object they wish to grasp.
Vibrotactile feedback is delivered to the forearm where the prosthesis is worn to facilitate the aiming.
Four vibrotactors provided tactile cues, guiding the user to align the optical $z$-axis with the target object (full details provided in Sec. \ref{sec:feedback} below).

Once the user correctly aims at the target object, the system begins processing the depth data to fit the three geometric models.
This step is crucial for reducing system latency, as ten frames are used to improve the accuracy of shape inference.
The target grasp size and wrist orientation values are determined using the median of a 10-frame moving window.
When the object is successfully reconstructed, all vibrotactors pulse simultaneously, signaling to the user that the prosthesis is ready and waiting for a command to preshape.
The user can then send an “auto-preshape" command to the prosthesis by performing a quick wrist flexion.
In response, the prosthesis adjusts its wrist orientation and grasp size according to the estimated target values.
If the user is not satisfied with the automatically selected hand configuration, they can aim again at the same object or a different object.

If the selected hand configuration is appropriate, the user can proceed to close the hand.
To do this, they would take over control by performing a quick wrist extension movement.
The prosthesis then moves proportionally to the recorded surface electromyography (sEMG) signals (direct control) and switches back to idle mode once the object is released.

\subsubsection{Vibrotactile Feedback for Aiming Guidance} \label{sec:feedback}

\begin{figure*}[t!] 
\centerline{\includegraphics[width=5in]{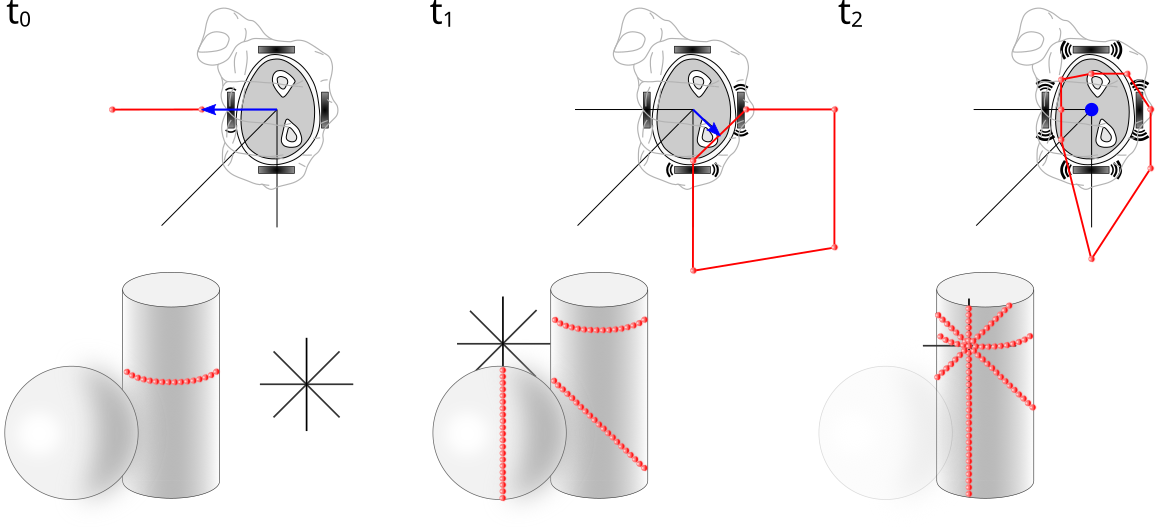}}
\caption{Illustration of the vibrotactile feedback used for aiming guidance. Four tactors placed on the forearm indicated the direction the user should move to position the optical $z$-axis “within" the object. The reference frame represents the sensor, whereas the user's aiming direction is given by the black crosses. The red polygons and blue arrows represent the convex hull of the combined laser scans in the $xOy$-plane of the sensor and the shortest-distance vector to that hull, respectively. At time $t_{0}$ the user is approaching the cylindrical object from the left (the right tactor pulses at low intensity). Then he/she slightly overshoots the aiming direction at time $t_{1}$, capturing thereby the data points from both objects (right and down tactors pulse at higher amplitude). Finally, the user corrects the aiming direction at time $t_{2}$ by moving slowly to the left, and once the optical $z$-axis “pierces" through the object (all four tactors pulse simultaneously at the maximum amplitude).}
\label{fig:tactors}
\end{figure*}

Fig. \ref{fig:tactors} shows the placement of the four tactors around the forearm.
This vibrotactile feedback strategy was implemented to improve the users' aiming accuracy, given the minimal amount of depth data available while operating the system.
The tactors begin vibrating as soon as the system detects an object, with the active tactor(s) indicating the direction in which the user should move the prosthesis, while the vibration intensity conveys the distance of the optical $z$-axis from the object.
Object detection and computation of the vibrotactile cues are performed using the raw depth data from the laser scans, prior to fitting the geometric models, as explained below.

When one or more depth points are detected within the capture volume, the aiming guidance procedure is triggered.
To simplify the problem, all points are projected onto the $xOy$-plane by setting their $z$-coordinate to zero.
The corresponding 2-D convex hull is then recursively computed using the QuickHull algorithm \cite{oRourke1998geometry}.
The norm and orientation of the shortest distance vector to the hull are used to cue the user toward the object(s).
This vector is normalized to the maximum width of the capturing volume and converted to polar coordinates.

The vibration signal is then conveyed to the user as a continuous signal, with the amplitude inversely proportional to the distance to the object: that is, the amplitude increases as the user approaches the object (Fig. \ref{fig:tactors}-A and \ref{fig:tactors}-B).
Active tactors are determined by the direction of the shortest-distance vector; If the orientation of the vector coincides with a given tactor, only that single tactor is activated, cueing the user to move left, right, up, or down.
In contrast, if the vector is between two tactors, both are activated, with weighted amplitudes depending on the projections of the vector, signaling to the user to move diagonally.

As soon as the user aims at the object (Fig. \ref{fig:tactors}-C) — that is, when the optical $z$-axis of the sensor “pierces" through the object — and the object is successfully fitted to one of the three geometric models, the aiming guidance cues switch to an intermittent buzzing pattern, with all the vibrotactor channels pulsating simultaneously.
Upon receiving this cue, the user can “trigger" the system to automatically move the prosthetic hand into the grasping configuration for the detected object, as explained in the previous subsection.

\begin{figure*}[t!] 
\centerline{\includegraphics[width=5in]{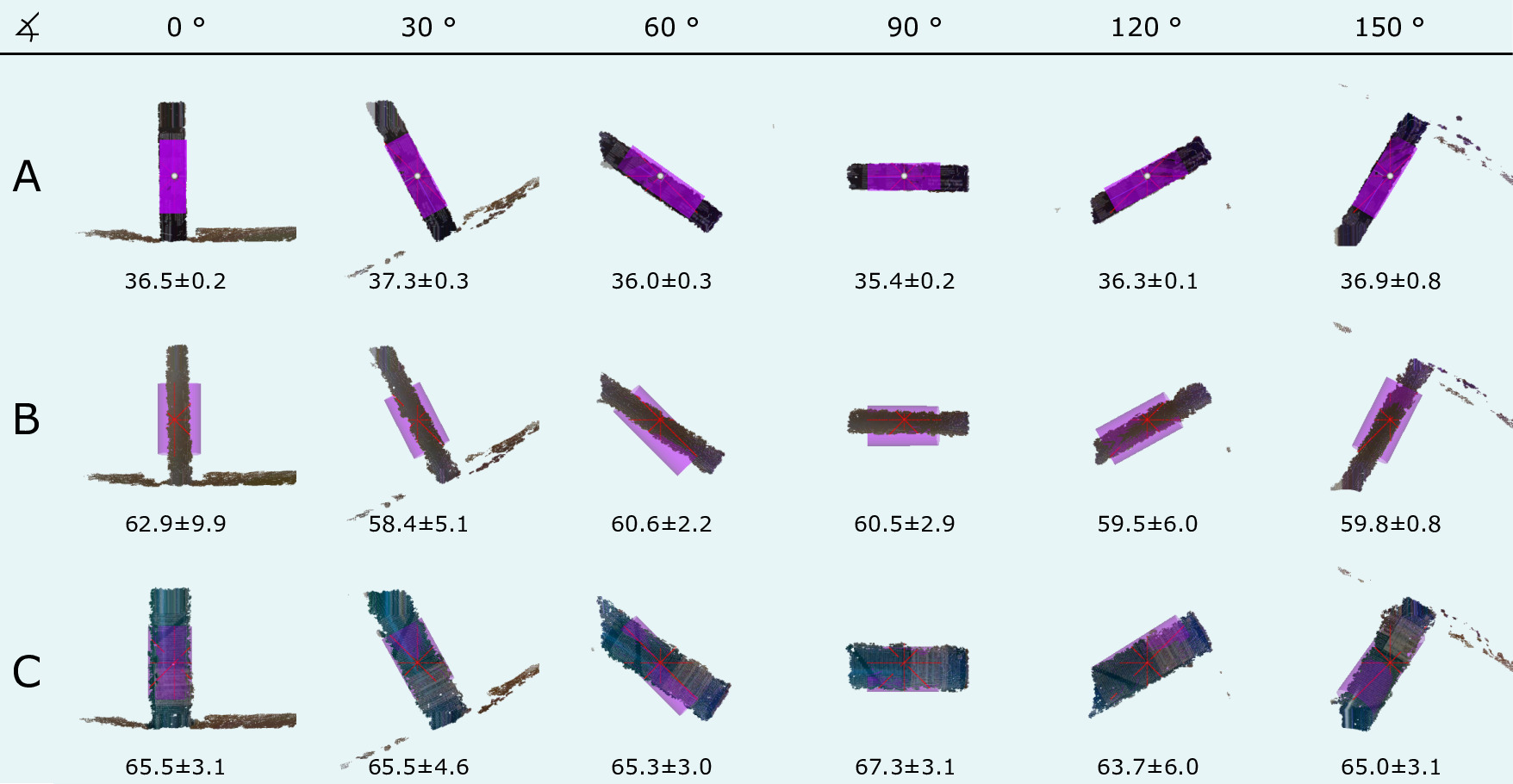}}
\caption{Rotational testing of the laser-scan-based reconstructions across six angles, in increments of 30\degree, for a rectangular prism with 40 mm width (A) and two cylinders with 40 mm (B) and 75 mm (C) diameter (all grasp estimation statistics are presented in mm). While the grasp size estimation for the 40 mm width rectangular prism (A) is correct, the reconstruction of its cylindrical counterpart (B) results in an overestimated aperture. This limitation is overcome when more depth data are available, as seen for the correct reconstruction of the 75 mm diameter cylinder (C).}
\label{fig:rotation}
\end{figure*}

\subsection{Performance Evaluation}

\subsubsection{Equipment}
A Michelangelo prosthetic left hand (Ottobock, Duderstadt, Germany) was used to test the proposed semi-autonomous control.
The prosthesis integrates two degrees of freedom (DoFs): wrist rotation and grasping with two different grips (palmar and lateral).
However, in this study, only wrist orientation and palmar grasp aperture were adjustable.
Two active/dry electrodes (Model 13E200=50, Ottobock, Duderstadt, Germany) were used to record pre-amplified and rectified double-differential sEMG signals.
The prosthesis was worn on the left forearm of able-bodied participants using a custom splint, while the electrodes were placed on the right forearm due to the lack of available space around the socket.
Importantly, this is not expected to impact the performance in any way since the myoelectric control was simple and robust, (2-channels of sEMG) and as such, unlikely to be affected substantially by the placement.
In fact, this is a general advantage of semi-automatic control approaches, where the automatic control decreases the impact of the individual differences and skills in volitional control.

The prosthesis was connected to a laptop (Intel\textsuperscript{\textregistered} Core\texttrademark{ }i7-8665U CPU @1.90GHz, 2.1GHz, 4 Cores, 8 Logical Processors, with 32 GB RAM) via a Bluetooth link.
The prosthesis sensor data and sEMG signals were sampled by the internal prosthesis controller at a frame rate of 100 Hz and transmitted to the laptop, which computed velocity control commands and sent them back to the prosthesis.

An active infrared stereo camera (Realsense{\texttrademark} D435i, Intel, US) was mounted dorsally on the hand.
The captured depth volume was defined as a virtual cropping box with dimensions of 100{x}100{x}100 mm, positioned 150 mm in front of the sensor.
The corresponding depth data stream, with a resolution of 424{x}240 px at 90 fps, was transmitted to the laptop via a two-meter active USB-A 3.0 extension cable (DELTACO Prime, SweDeltaco AB, Stockholm, Sweden).
The four concurrent laser lines were then extracted from these point cloud data, by subsampling the data such that the four scan lines were 45\degree~apart from each other.
This has effectively emulated the output of a real laser scanner with four scanning rays.
A tolerance of 1 mm was applied to determine whether a given data point belonged to a specific laser scan; points outside this tolerance were discarded.
The data processing pipeline was implemented in C++ using the open-source Point Cloud Library (PCL) v1.11.1 \cite{Rusu_ICRA2011_PCL}.
Custom code was developed to add an ellipse model for RANSAC fitting to the PCL (see Appendix \ref{appendix_ellipse}).

Vibrotactile sensory feedback was also provided to participants on the same forearm carrying the prosthesis to assist with aiming towards target objects while using the laser scan prototype.
This feedback system (C2 tactors, Engineering Acoustics Inc, Casselberry, FL, USA) was connected directly to the laptop via a USB cable and operated as described in Sec. \ref{sec:feedback}.
The tactors pulsing intensity was adjusted and checked for each individual participant prior to the experiment.
Note that, for clinical applications, all components would be embedded within the prosthesis socket; however, such integration was beyond the scope of the present work.
Importantly, a recent study \cite{maravic2024feeby} demonstrated that vibrotactile stimulation would not interfere with sEMG recording even when both are placed within the socket.

\subsubsection{Experimental Protocol}
\begin{figure*}[t!]
\centerline{\includegraphics[width=5.0in]{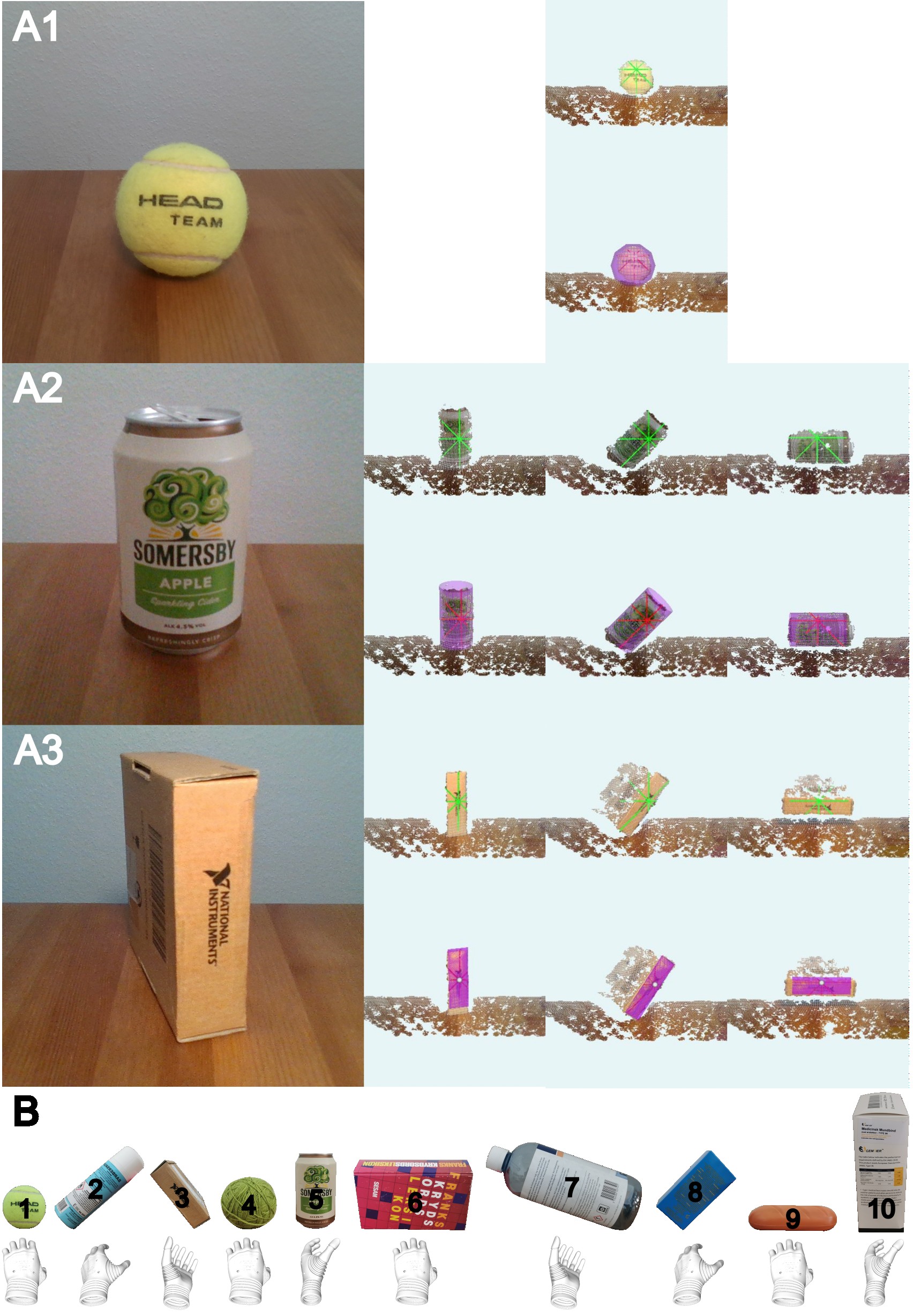}}
\caption{(A) Reconstruction of target objects that resembled different 3-D geometric shapes (sphere, cylinder, and cuboid). The laser scanners are represented by four green lines in a ”star” configuration. These scans are used by the system to interpret the 2-D shape of the partial object contours, which are then used for the full reconstruction of each object. The reconstructed 3-D geometric models are shown in purple superimposed on each object. (B) The ten object sequence used in the experimental protocol presenting each object in its respective orientation.}
\label{fig:perform}
\end{figure*}

Ten able-bodied adults (24.5$\pm$3.9 years old) were recruited and the experimental protocol was approved by the Research Ethics Committee for North Jutland (N-20190036).
All participants provided written informed consent prior to participation.
The experiment involved a series of pick-and-place tasks using a set of ten objects selected to include the most common object shapes and orientations that the prosthesis users need to handle in daily life: two spherical, four cylindrical, and four cuboid objects, with grasping sizes ranging from 35 mm to 85 mm.
The four cylindrical and cuboid objects were placed in various orientations — upright, lying down, tilted to the left, and tilted to the right — to challenge the system’s adaptability.
A fixed object presentation sequence was defined for all participants, which required them to rotate the wrist and adjust the hand aperture of the prosthesis for each upcoming object, thus preventing consecutive grasps with similar orientations.
This sequence, as illustrated in Fig. \ref{fig:perform} (bottom), was repeated five times (blocks).

The performance of this laser scanning approach was compared against a benchmark system \cite{castro2022continuous} that processed full point clouds.
The only modifications from that study was that hand preshaping was not continuous but instead triggered by the user, to match the approach used in the present prototype, and the hand could only close in palmar grasp (lateral grasp was disabled).
Vibrotactile feedback was not provided to the participants for the benchmark system given that it only requires aiming in the vicinity of the object, eliminating thus the need for precise aiming.
Both systems were tested by the participants, on the same day, following the same experimental protocol.
Specifically, half of the participants (randomly picked) started with the benchmark system and then tested the prototype, and vice-versa for the other half.
As explained above, the objects were always presented in the same order.
The experiment lasted two hours per participant and none of them had prior experience with either of the two controllers.

An experimenter introduced the participants to both systems.
To maximize the effectiveness of the vibrotactile feedback, the participants were instructed to approach the objects from the side as soon as they perceived the tactors vibrating.
Each participant was allowed a maximum of three tryouts to familiarize themselves with these indications.
During the experiment, the participants waited for a voice command from the experimenter to initiate the following steps: 1) move the prosthesis towards the target object, 2) trigger the automatic controller to preshape the hand, 3) take control of the prosthesis, and 4) manually close the hand to grasp the object and move it to a marked area (0.5 m from the initial position).
If the object slipped after the participant took control, the trial was repeated.

The primary outcome measure for assessing performance was the Time to Accomplish the Task (TAT), defined as the time interval from when the participant began moving towards the object until the object was released in the designated “drop" zone.
TAT was chosen as an outcome measure because time is employed as the main performance measure in many other tests for assessing prosthesis use, e.g. the Southampton Hand Assessment Procedure (SHAP) test \cite{light2002establishing} and the Box and Blocks test \cite{mathiowetz1985adult}.
Note that, if the automatically adjusted hand configuration was incorrect (e.g., an aperture that was not wide enough or a clearly wrong orientation), the subjects had to step back and trigger the automatic control again, which resulted in longer TAT.
Additionally, to evaluate the performance of the computer vision pipeline, we calculated the reconstruction accuracy as the percentage of objects whose models were correctly estimated, along with the error in estimating the object size for both the novel controller prototype and the benchmark.

\subsubsection{Data Analyses}
The median TAT across objects was calculated for each participant in each block.
A Shapiro-Wilk test revealed that the data did not follow a normal distribution; consequently, a Friedman test was employed to evaluate whether the performance of the two systems varied across blocks.
The posthoc pairwise comparisons were performed using Wilcoxon signed rank tests (statistical significance of $p\leq0.05$), with Bonferroni correction.
Additionally, a Wilcoxon signed rank test was also used to compare the performance between the two systems within each block.
The results presented in Sec. \ref{sec:results} are reported as M\{IQR\}, where M represents the median and IQR represents the interquartile range.

\begin{figure*}[t!]
\centerline{\includegraphics[width=5in]{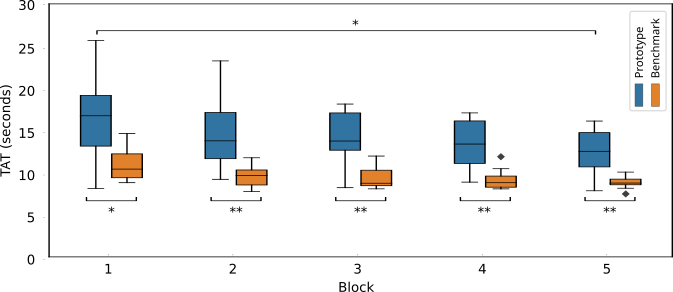}}
\caption{Summary results for the Time to Accomplish the Task (TAT) for both systems across blocks. The box plots indicate median, interquartile range, and min and max values; asterisks show statistically significant differences (*$p\leq0.05$, **$p\leq0.01$).}
\label{fig:stats}
\end{figure*}

\subsection{Testing of Irregularly Shaped Objects}
The reconstruction capabilities of our system were further evaluated using a supplementary set of objects, whose shapes significantly deviated from the geometric models typically modeled.
The goal was to stress-test the prototype, as many objects encountered in daily life cannot be perfectly represented by the aforementioned geometric shapes.
This test set consisted of ten objects, including a prism-like box (hexagonal prism), a bar of soap (spherocuboid), a TV remote with buttons (irregularly surfaced cuboid), a horse toy (potentially an ellipsoid, assuming aiming at the torso), a sweet potato (irregular ellipsoid), a banana (curved cylinder), a pear (sphere/ovoid), the cap of a small jar (disc, approached both frontally and from above), a conical mug (cylinder with varying diameter), and a frying pan handle (cuboid with varying width).

The success criteria for evaluating whether the prototype would be able to correctly align with and open for picking these objects was set at a maximum estimation error of $\pm20$ mm from the actual size and of $\pm10$\degree~from the actual orientation (with standard deviation of less than 15\degree).

\section{Results} \label{sec:results}

Fig. \ref{fig:rotation} presents the capabilities of the proposed laser-scan approach when reconstructing basic objects, such as a rectangular prism or a cylinder with two distinct diameters, for six different angles in increments of 30 \degree.
The trivial evaluation of a sphere is not presented since the object is rotation invariant.
On average, the estimations of the true 40 mm grasp size were $36.4\pm0.3$ mm (9\% error) and $60.3\pm4.5$ mm (51\% error, an example of a failed estimation) for the rectangular prism and narrower cylinder, respectively.
The grasp size for the wider cylinder with a cross-sectional diameter of 75 mm was estimated to be, on average, $65.4\pm3.8$ mm (13\% error).
Note that the size estimation do not change much with orientation.

The testing of the reconstruction accuracy for the ten objects selected for the experimental assessment showed that, for every 100 samples, on an individual frame basis, the novel controller prototype successfully reconstructed 96.5\% spherical objects, 95.3\% cylindrical objects, 89\% cuboid objects, compared to 98\%, 93\% and 100\%, for the benchmark approach, respectively.
The subsequent reconstruction accuracy obtained with the moving window of size ten was, for the prototype, 100\% on spherical objects, 97.5\% on cylindrical objects, 100\% on cuboid objects, whereas the benchmark approach reached a 100\% accuracy for every object type.
The mean absolute error in estimating the object size was $8\pm5$ mm ($13\pm9$ \% error) for the prototype and $4\pm3$ mm ($6\pm4$ \% error) for the benchmark system.
The mean absolute estimation error in object orientation was $2\pm3$\degree~for both systems.

The computational time for both systems was comparable. On average, the prototype and benchmark took 138 ms (7 Hz) and 75 ms (13 Hz) per reconstruction, with refresh rates of [4, 17] Hz and [9, 18] Hz, respectively.
For a resolution of 424{x}240 px, the proposed method uses 700$\pm$300 data points, while the benchmark uses 5300$\pm$2100 data points.

Representative examples of object reconstructions from the laser scans for some objects used in the experiment are provided in Fig. \ref{fig:perform}, top.
The figure shows the scan lines and the estimated object models.
The cuboid and cylindrical objects are placed in three different orientations to demonstrate that the system can estimate object parameters (model, size, and orientation) regardless of the current prosthesis configuration or object posture.
In the case of the cylindrical object, the laser scans generated when the object was placed vertically and horizontally produced a line, a circle, and two ellipses.
In contrast, for the tilted cylinder, the scans are obtained by fitting four ellipse models.
The cuboid object, however, was reconstructed using four line models regardless of orientation.
The lines were placed across its frontal face to which the participant aimed at, but he/she could have selected any other face to be the target for grasping (e.g., approaching the object from the top side).

Fig. \ref{fig:stats} presents the TAT across blocks for both approaches.
The median TAT decreased for both approaches, from 16.7\{6.0\}s to 12.5\{4.0\}s for the prototype (25\% of reduction) and from 10.4\{2.9\}s to 8.8\{0.7\}s for the benchmark (16\% reduction).
The statistically significant improvement was, however, found only for the laser scanning approach, where the TAT in the fifth block was significantly smaller ($p\leq0.05$) than the TAT in the first block.
The benchmark method consistently outperformed the laser scanning, as the participants were significantly faster with that system in each block.
Nevertheless, the difference in TAT between the two systems consistently reduced across blocks, from 6.3 s in the first to 3.7 s in the last block, and this reduction was statistically significant ($p\leq0.05$ or $p\leq0.01$).
As a side note, a total of 13 and 17 trials (beyond all 1000 recorded) were deemed as failed for the prototype and the benchmark, respectively.
The two objects with the largest amount of failed attempts across both modalities were objects 3 and 7.

\begin{figure*}[t!] 
\centerline{\includegraphics[width=5in]{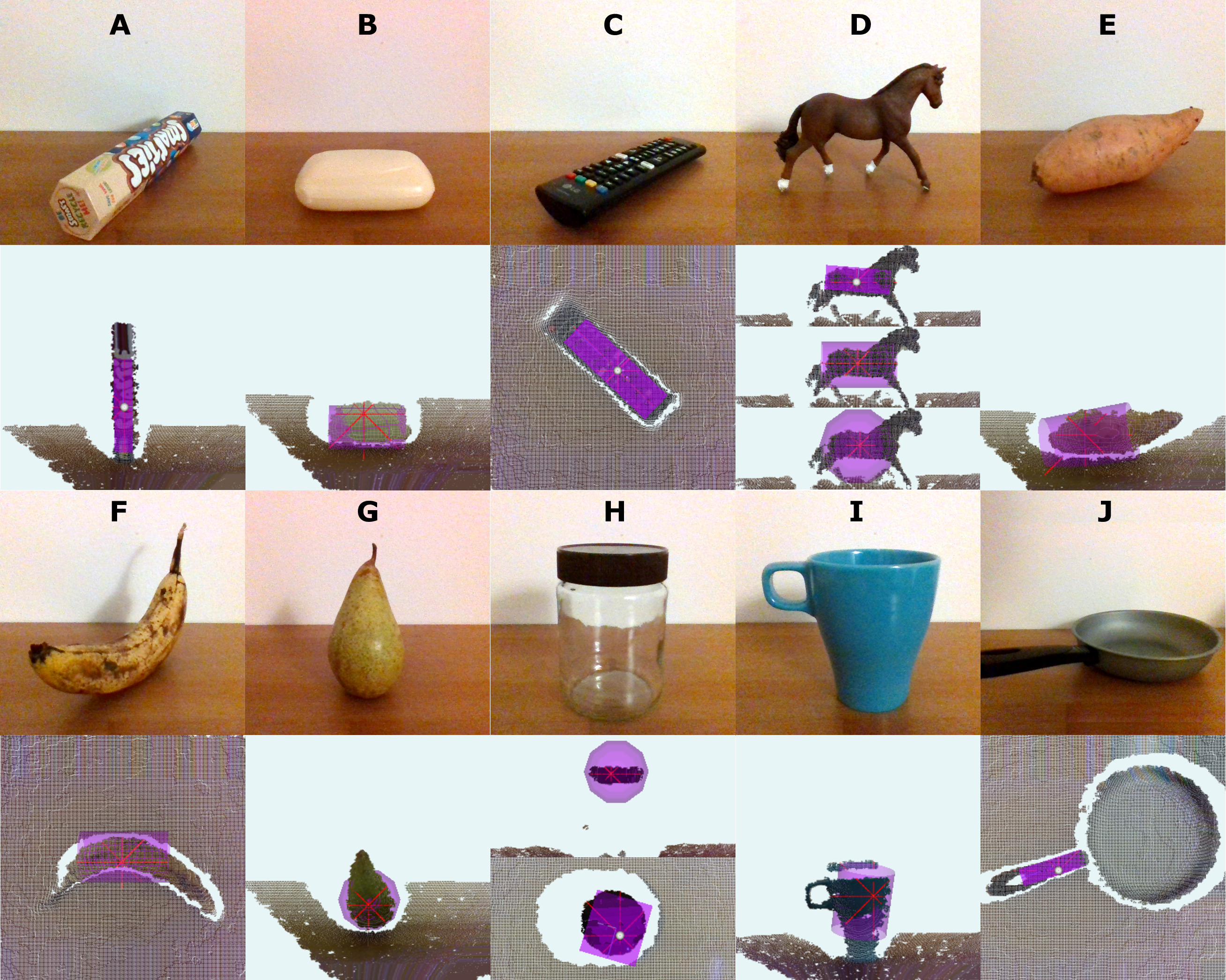}}
\caption{Laser scan-line-based reconstruction of the objects with irregular shapes: a (A) prism-like box (hexagonal prism); (B) a bar of soap (spherocuboid); (C) a TV-remote with buttons (irregularly surfaced cuboid); (D) a horse toy (potentially ellipsoid, assuming aiming at the torso); (E) an irregular sweet potato (an irregular ellipsoid); (F) a banana (curved cylinder); (G) a pear (sphere/ovoid); (H) the cap of a small jar (a disk, seen both frontally and from above); (I) a conic mug (cylinder with varying diameter); (J) and a frying pan handle (cuboid with varying width).}
\label{fig:exotic}
\end{figure*}

The reconstruction results of the objects with irregular shapes are presented in Fig. \ref{fig:exotic}.
The respective statistics are presented in Table~S1. 
The system successfully inferred the expected 3-D model shape approximation for most objects except for the (D) horse toy, the (F) banana, and the (H) cap of a small jar.
While trying to fit these objects, there was a significant spread in object reconstruction, i.e. shape inference, which led to erroneous grasp size and orientation estimations with a high degree of variability.
Besides, when approaching the disk-shaped object (D) from above, the system failed to fit a 3-D model most of the time.
Overall, the expected grasp size was slightly overestimated with respect to the actual object width.

\section{Discussion}
This study proposed and evaluated a new semi-autonomous prosthesis controller using depth sensing from a hand-mounted sensor.
Unlike previous works that relied on heavy processing of full point clouds \cite{markovic2014stereo,markovic2015sensor,mouchoux2021decreases,castro2022continuous}, our approach utilizes minimal data from multiple laser scans.
We explored the effectiveness of using between one to four scan lines, concluding that four lines are necessary for reliable reconstruction in all relevant cases.
A novel method was developed to reconstruct object shape, size, and orientation from these four scans using simple geometry and the fitting of 2-D geometric models — such as circles, ellipses, and lines — instead of explicit 3-D models as seen in prior art \cite{markovic2014stereo,markovic2015sensor,mouchoux2021decreases,castro2022continuous}.

The initial evaluation of the prototype on different objects (Fig. \ref{fig:rotation}), including a rectangular prism and two cylinders, demonstrated that the system could estimate an appropriate grasp size, regardless of the object's or wrist's orientation.
Nonetheless, it struggled with narrower cylinders, often overestimating the required grasp size.
This occurred because fewer data points were captured given the small circumference arc of their cross-section, which lead to poorly RANSAC fitting.
While this could be seen as a limitation, a slightly larger hand aperture is preferable to a narrower one, which would require an additional command to reopen and correct the hand.
A narrow cylindrical object (see object 9 in Fig. \ref{fig:perform}) was also included in the functional assessment.
While the objects in Fig. \ref{fig:perform} are representative of the most common grasping targets and hence most relevant functionally, the system was additionally stress tested by attempting to reconstruct the irregularly shaped objects (Fig. \ref{fig:exotic}).
Despite the complexity of objects like pears, potatoes, mugs, and large handles, the system was able to accommodate them successfully.
Importantly, although the system was tested on a limited set of objects, our results are more general.
The novel approach uses depth information to reconstruct the object properties and decide hand preshape, and hence, if the approach works for a specific object type (e.g., bottle), it can also handle other objects of the same type (e.g., other bottles). 

The functional assessment showed that the system could accurately reconstruct a set of daily life objects (Fig. \ref{fig:perform}) and allowed participants to complete pick-and-place tasks (Fig. \ref{fig:stats}).
Contrary to our hypothesis, participants were slightly slower using the novel controller prototype compared to the benchmark system, though the difference was small.
This gap in TAT performance decreased over time, with only a ~4 s difference after 50 grasp attempts, suggesting it could diminish further with more practice.
The tendency for this further decrease as well as for the time necessary to reach a plateau performance, needs to be investigated in future work.
The lower performance of the new prototype was mainly due to the need for precise object aiming, as full reconstruction only occurred when the optical $z$-axis “pierced" the object.
In contrast, the benchmark system was more forgiving, requiring only that the object be within the general area in front of the prosthesis \cite{castro2022continuous}.
Some participants also struggled when grasping two objects positioned in a way that required the prosthetic hand to move underneath, as their own hand obstructed their view.
This issue stemmed from the lower positioning of the prosthetic hand on the forearm split but would likely not occur with a prosthetic hand mounted on a proper socket.

Vibrotactile feedback was introduced to assist with aiming in the novel controller prototype.
Since the tactors were placed around the forearm, as shown in Fig. \ref{fig:control}, as far as possible from each other, the participants could easily discriminate their activation.
In addition, the maximum intensity of each tactor was adjusted for each participant individually so that they elicited clear sensations that were similar in intensity across tactors.
Finally, it is important to emphasize that to successfully use the system, the participants did not need to interpret this feedback precisely – instead, the role of the feedback was to prompt the subject to point the prosthesis in the general direction of the object as shown in Fig. \ref{fig:tactors}.
Although we trained the participants to aim before commencing with the experiments, the TAT of the proposed system remained higher compared to the benchmark.
Sensory feedback has been used in vision-based, semi-autonomous prosthesis control systems before \cite{markovic2014stereo}, but typically through visual displays such as AR glasses \cite{markovic2014stereo, mouchoux2021decreases} or blinking LEDs \cite{castro2022hybrid}), rather than tactile feedback as in this study.
Tactile feedback offers advantages by integrating into the prosthesis and avoiding additional visual load, which can already be high when using a prosthesis \cite{markovic2014stereo, mouchoux2021decreases}.
An alternative for simplifying aiming could be a guiding laser pointer, as in Dosen et al. \cite{dovsen2010cognitive}, but tactile feedback remains more discreet and non-intrusive.
Nevertheless, it remains to be investigated how such feedback would be accepted by the users.
In principle, if the users find it intrusive, the feedback could be employed only initially for the training and then deactivated.
Alternatively, the feedback could be made more attractive by using it to provide other variables (e.g., grasping force) during manual control (e.g., after the object was grasped).

The proposed prototype system used seven times less data compared to the benchmark.
Furthermore, the amount of data increases linearly for the prototype with higher resolution, compared to quadratically for the benchmark.
The same execution speed for both methods was likely because the data was processed using a high-performance laboratory laptop, whereas the performance difference would show up once the data processing pipeline is optimized to be deployed to embedded hardware.
In the latter case, even if the execution speed would not decrease substantially, the fact that significantly fewer data points need to be stored, transmitted and processed would still be a rather important advantage of the novel method.
The focus of the present work was on proving the feasibility of reconstructing objects with minimal data, while optimizing the pipeline to fully leverage this reduced data and simpler processing, and ultimately deploying the processing to an embedded system remains future work.
Despite this, the performance of object reconstruction was generally rather high for both approaches.
The same holds for the orientation, where the maximum error was 5\degree.
Importantly, even much larger orientation errors could be corrected using compensatory motions (but it is biomechanically beneficial to avoid those as much as possible).
However, using real higher-quality laser scanners is expected to greatly improve estimation precision.

While the laser scans were obtained from a depth camera for testing flexibility, the final solution aims to incorporate a custom-made light detection and ranging (LiDAR) or an indirect time-of-flight (ToF) sensor, such as a photonic mixer device (PMD) sensor.
This transition would make the system more compact and enhance scan quality.
The sensor could be integrated into the hand or socket and potentially simplified to fewer than four scans, though this would limit functionality and assistance for the prosthesis user (Sec. S.I.).

Although crossing lasers have been previously used for 3-D reconstruction with projected light \cite{Furukawa2009, he2017cross}, laser interference presents a challenge in ToF technology \cite{mcmanamon2012ladar}.
This ToF technology has been widely used in other robotics areas such as localization and mapping \cite{kallasi2016keypoint}, automated inspection \cite{yamada2013road}, vehicle counting \cite{mogelmose2016wheels} or forestry \cite{jutila2007forest}.
Yet, mutual interference primarily occurs in co-planar setups \cite{kim2015mutual}, which is not applicable to our proposed implementation.
Interference can be direct, when two scanners face each other, or indirect, when light from one emitter reflects off an object and scatters toward a second receiver \cite{kim2015mutual}.
The scanning system could utilize a sensor that emits in multiple planes with a single detector.
To avoid indirect interference, different laser scans could be pulsed alternately, further reducing light scattering effects, especially since the scanners are not co-planar \cite{kim2015mutual}.
Most interference issues stem from multiple laser scanners operating at the same wavelength, typically around 905 nm \cite{li2020driving}.
By using different near-infrared wavelengths, we can mitigate indirect interference \cite{sun2019siphotonics}.

The present work has some limitations.
Our method fails to reconstruct objects with diameters below 40 mm, especially cylinders and spheres, due to fewer points captured, often leading to line models being fitted instead of circular or elliptical.
Nevertheless, this will ultimately depend on the quality of laser scanners used to implement the prototype in the future work.
The same applies to highly irregular shapes like bananas or small toys, since the system is limited to three geometric models.
These reconstruction errors led to unstable grasp aperture and wrist orientation, resulting in continuous jerky movements, making them essentially ungraspable, unless the user takes over the control and configures the hand manually.
Therefore, in this case, the system would simply fall back to the conventional control.
A similar issue occurred with flat cylinders (height $\leq 40$ mm, like a jar cap), where too few data points were captured, especially from frontal views, leading to poor cylinder fitting - less than twice that of a sphere fitting (Sec. \ref{sec:sysoperation}).
Approaching these objects from above also yielded unsuccessful reconstructions, as flat disks do not provide enough data for the system to compute a reliable principal direction vector - no three circumference points separated by 45\degree~can ever be co-linear, according to Algorithm \ref{alg:seeds}.

The tactile feedback scheme also showed a limitation in distinguishing between two closely positioned objects.
If both objects are within the depth sensor's field-of-view and their data points are detected at the edges of the acquisition volume, aiming at the space between them could cause the convex hull to be mistaken for a single object.
This would lead to simultaneous vibration of all tactors.
However, this is unlikely given the narrow 100 $mm^{3}$ depth acquisition volume.

The overall cognitive burden was not assessed, as prior studies \cite{markovic2015sensor,mouchoux2021decreases} and the benchmark \cite{castro2022continuous} have already shown the advantages of semi-autonomous control over manual control, especially when the prosthesis is more complex (2+ DoFs) — this would not be beneficial for a simple open and close gripper prosthesis.
In addition, despite aiming was effective, it must be investigated how the users would react to this extra step in prosthesis use.
The feedback was added to facilitate this step, but this introduces an overhead as the participants need to interpret vibrotactile sensations.
Importantly, we assume that with practice and integration fo the lasers into the hand, the aiming could become more routine, and less dependent on the feedback.
Testing on real amputees is pending the design of a custom socket with a fully integrated system, which is a configuration relevant to clinical applications.
Similarly, the assessment of system robustness to different environments and lightning conditions (e.g., use in and out of the lab) will be performed once the integrated system is available, since these aspects will critically depend on the characteristics of the real scanners.

The experiments demonstrated that the developed prototype reliably enabled participants to handle various objects with different shapes, sizes, and orientations.
The fact that both the novel system and the benchmark could be tested in the same session, and after only a brief explanation of how the system works, illustrates the benefits of semi-autonomous control.

This work marks an important step towards the clinical applicability of semi-autonomous prosthesis control approaches.
By minimizing the data needed to estimate object properties, a simpler, smaller, and more cost-effective sensor could be utilized, reducing power consumption and the memory and computational resources required for data storage and processing.
These factors are crucial, given the strict size, power, and computational constraints of wearable prostheses.
Creating a fully self-contained system could involve using a combination of laser scanners instead of a depth camera.
The present study demonstrated the feasibility of the approach, and the next step is to test the system by placing real laser scanners on the prosthesis.

\appendix[Implementing an Ellipse3D for RANSAC fitting] \label{appendix_ellipse}
A 3-D ellipse model based sample consensus algorithm was implemented and merged to the PCL \cite{Rusu_ICRA2011_PCL} repository. Code: \url{https://github.com/mnobrecastro/pcl-ellipse-fitting}.

The iterative process of RANSAC \cite{fischler1981random} using a 3-D circle or ellipse geometric model involves the following steps:
1) randomly select k (k=3 or k=6, respectively) inlier points from the set to initialize the algorithm;
2) define the normal vector to the average plane containing those points and pick one of them;
3) calculate the projection of all k points onto that plane;
4) fit the geometric model;
5) measure the distances from the remaining points in the set to the geometric model;
6) evaluate whether each of those points is an inlier or outlier based on a preset distance threshold.

For an ellipse, Step 4) can be accomplished through a linear least-squares ellipse fitting during Step 4).
This method presented by Fitzgibbon et al. \cite{Fitzgibbon1999} can handle noisy data and provides the six parameters of the conic equation of an ellipse by solving a constrained generalized eigen system.
The model is then converted to a parametric model with 11 parameters: the center point (three); the semi-major/minor distances along the local x- and y-axis of the ellipse (two); the normal to the plane containing the ellipse (three); and the local x-axis of the ellipse (three).
Afterwards, in our implementation, Step 5) first identifies to which ellipse quadrant each point relatively belongs.
Then, it performs a golden-section line search \cite{arora2004introduction} on the ellipse points belonging to that quadrant in order to find the shortest distance to the ellipse.

\ifCLASSOPTIONcaptionsoff
  \newpage
\fi

\bibliographystyle{IEEEtran}
\bibliography{References}

\end{document}